\documentclass[runningheads]{llncs}
\usepackage[T1]{fontenc}
\usepackage{amsmath,amssymb}
\usepackage{graphicx}
\usepackage{hyperref}
\begin{document}

\title{Towards Value-Constrained Credit Assignment in Fully Delegated AI Cooperatives}
\titlerunning{Value-Constrained Credit in Delegated AI Co-ops}

\author{Young Yoon\inst{1}\orcidID{0000-0002-5249-2823} \and
Jimin Kim\inst{1}\orcidID{0009-0007-6078-183X} \and
Soyeon Park\inst{1}\orcidID{0009-0005-7401-8902}}

\authorrunning{Y. Yoon et al.}

\institute{
Hongik University, Seoul, Republic of Korea \\
\email{young.yoon@hongik.ac.kr}\\
\url{https://dina.hongik.ac.kr/}
}

\maketitle

\begin{abstract}
We propose a framework for reward allocation in fully delegated AI cooperatives where humans are represented by agents that contribute data and participate in model updates under heterogeneous value constraints. The key idea is to credit only those updates that remain admissible after screening them against each principal's value profile. We formulate value-conditioned gradient filtering, online marginal contribution signals, and cumulative revenue settlement within a traversal learning (TL) substrate. TL is especially attractive here because it performs decentralized backpropagation without the quality loss associated with aggregation-centric distributed learning and, we argue, offers a finer attribution substrate than FedAvg-style federated learning by preserving explicit traversal and gradient paths. The framework is positioned against data valuation, federated contribution estimation, personalized federated learning, and pluralistic alignment.
\end{abstract}

\section{Introduction}
AI cooperatives pool data and computation from multiple members to produce shared AI services. In a fully delegated cooperative, each member is represented by an autonomous agent that participates in learning on the member's behalf. Two requirements then arise simultaneously: learning must respect the member's value constraints, and the economic benefits of the resulting service must be distributed according to actual contribution.

These requirements are coupled. If one member disallows military deployment, another rejects insurance risk scoring, and another requires fairness across demographic groups, then not every update induced by their data should count equally toward a common model. A reward rule that ignores such constraints may compensate members for updates they would not endorse, while a value-aware learning rule with contribution accounting offers sustainable incentive structure.

Traversal learning (TL) is a useful substrate for this setting. TL is a form of decentralized backpropagation that avoids the degradation seen in FL, SL, and SplitFed-style schemes by preserving centralized-learning-style gradient computation across distributed nodes \cite{batbaatar2025tl}. Relative to FedAvg-style FL \cite{mcmahan2017communication}, TL also makes attribution structurally clearer: because learning proceeds through explicit traversal and orchestrated gradient flow rather than only aggregated client updates, the path from local data exposure to admissible update to cooperative effect can be tracked more transparently. This is particularly important when contribution must be measured only after checking value admissibility.

The proposed pipeline has four steps: a principal delegates data and a value profile to an agent; the agent computes a local update and filters it through that value profile; the cooperative estimates the marginal contribution of the admissible update to shared validation quality; and revenue is allocated according to these contribution scores.

\section{Problem Setting}
%Consider an AI cooperative built on electronic medical record (EMR) data. Members may all contribute clinically useful records while differing sharply on allowed downstream use: some prohibit military or surveillance applications, some reject insurance risk scoring, and some require fairness constraints before commercial deployment. 
The core problem is not only to learn from distributed data, but to decide which update directions are admissible for each member and how to reward the remaining, value-consistent contribution. Formally, agent $i$ holds local data $D_i$, local loss $L_i(\theta)$, and a value profile $v_i$. We use $L_{\mathrm{val}}(\theta)$ for cooperative validation accounting. A practical question is how the value profile should be represented at delegation time. Two broad precedents are useful. First, values may be given as explicit rule sets or constitutions, as in Constitutional AI \cite{bai2022constitutional}. Second, they may be learned from preference data or pairwise comparisons, as in preference-based reward modeling \cite{christiano2017preferences}. In this paper, $v_i$ is treated abstractly so that either representation can be used.

\section{Framework}
\subsection{Value-Conditioned Filtering}
At iteration $t$, agent $i$ computes a local gradient $\nabla L_i(\theta^t)$ and applies a value filter:
\begin{equation}
\tilde{\nabla L}_i(\theta^t)=\mathcal{F}_i(\nabla L_i(\theta^t),v_i).
\end{equation}
The filtered gradient $\tilde{\nabla L}_i$ is the admissible part of the update under the principal’s values. This defines pluralism at the gradient level: agents selectively admit or reject update directions rather than whole tasks.

The filter may take the form of rule-based constitutional checks \cite{bai2022constitutional}, feasible-set projection \cite{yang2020pcpo}, gradient modification \cite{yu2020pcgrad}, or a learned black-box admissibility model trained from preference data \cite{christiano2017preferences}.

\subsection{TL-Based Cooperative Update}
Using TL for training, we write the cooperative update abstractly as
\begin{equation}
\theta^{t+1}=\theta^t-\eta\sum_{i=1}^{n}\tilde{\nabla L}_i(\theta^t),
\end{equation}
where $\eta$ is the learning rate. This notation suppresses the internal traversal schedule but keeps the property we need: admissible updates are propagated through explicit agent-specific gradient paths rather than only post hoc aggregated client states. This distinction matters for attribution. In FL contribution estimation, one often has to reconstruct contribution from already aggregated client updates \cite{chen2024flce}. In TL, the traversal and orchestrated backpropagation structure keeps the update pathway more explicit. We therefore argue that TL supports a finer and more transparent accounting of value-consistent contribution than aggregation-centric FL.

\subsection{Marginal Contribution and Settlement}
We define the time-$t$ contribution of agent $i$ as the marginal improvement induced by its admissible update:
\begin{equation}
c_i^t=L_{\mathrm{val}}(\theta^t)-L_{\mathrm{val}}\bigl(\theta^t-\eta\tilde{\nabla L}_i(\theta^t)\bigr).
\end{equation}
This is a one-step counterfactual quantity: how much would validation improve if only agent $i$'s admissible update were applied at the current state? If the filter removes the update, then $\tilde{\nabla L}_i=0$ and $c_i^t=0$, so inadmissible directions cannot generate credit. The quantity $c_i^t$ should be read as an online credit signal rather than as the final payment basis. Settlement is performed over an accounting horizon $H$ by accumulating stepwise admissible contributions:
\begin{equation}
C_i(H)=\sum_{t\in H}\omega_t c_i^t,
\end{equation}
where $\omega_t$ is an optional weighting term that can discount repeated exposure or emphasize later checkpoints. Revenue shares are then allocated from cumulative contribution:
\begin{equation}
r_i(H)=\frac{(C_i(H))^\alpha}{\sum_{j=1}^{n}(C_j(H))^\alpha},\qquad \alpha\ge 1.
\end{equation}
If the cooperative generates total revenue $R(H)$ over the same horizon, the settlement payment is
\begin{equation}
p_i(H)=R(H)\,r_i(H).
\end{equation}
When $\alpha=1$, payment is proportional to cumulative contribution; larger $\alpha$ accentuate differences. Agents are rewarded only for updates that are both admissible and validation-improving.

\section{Related Work and Positioning}
Shapley-based data valuation and influence-based attribution quantify the value of data or updates with respect to model behavior \cite{shapley1953value,ghorbani2019data,koh2017understanding}. However, these approaches are not cooperative payment rules under heterogeneous value constraints. Contribution estimation and incentives in FL are already active research areas \cite{chen2024flce,murhekar2023incentives}. Yet those methods typically start from aggregation-based FL and ask how to estimate or pay for contribution after aggregation. Unlike prior approaches, our framework evaluates each update for value admissibility online within the training loop before its contribution is measured.

Personalized FL addresses heterogeneous client objectives, but primarily for predictive personalization rather than normative acceptability \cite{luo2023pgfed}. Pluralistic alignment highlights that serving diverse human values remains an open problem \cite{sorensen2024pluralistic}. Our framework complements that literature by moving pluralism into the training-time credit-allocation mechanism itself. Its novelty is therefore not a new attribution metric alone, but the integration of value-conditioned update filtering, TL-based attribution, online contribution signals, and cumulative cooperative settlement in one mechanism.

\section{Discussion and Conclusion}
The framework does not solve pluralism by collapsing heterogeneous values into one objective. Instead, it makes pluralism operational inside learning dynamics. Classical convergence to a single optimum need not hold because agents may admit different update directions; the more realistic expectation is stabilization within a value-constrained feasible region or compromise equilibrium. Importantly, even when global convergence is imperfect, the local credit signal remains computable at each step, so cumulative settlement over an accounting horizon is still well-defined.

For greater efficiency and privacy, a TL variant~\footnote{https://github.com/neouly-inc/TLplus} could place only a cut layer or shallow front block at each member agent, transmit safely masked or privatized activations to the orchestrator, and let the orchestrator execute the remaining backpropagation~\cite{batbaatar2026tlaccuracyprivacypreserving}. This direction is motivated by SplitNN-style partitioning \cite{vepakomma2018split} and recent privacy-preserving methods that protect smashed activations through masking or noise injection \cite{oh2024dpcutmixsl}. Such a design could preserve the attribution benefits of TL while reducing client-side cost and limiting information leakage from intermediate representations. We do not claim to outperform Shapley-style valuation in fairness or FL incentive mechanisms. Rather, we argue that TL offers a more transparent attribution substrate than aggregation-based FL, and that this substrate makes it possible to connect value admissibility, contribution accounting, and cooperative reward allocation in a single distributed-learning framework.

\subsubsection{\ackname}
    This work was supported by the MSIT (Ministry of Science and ICT), Korea under the ITRC (Information Technology Research Center) support program (IITP-2025-RS-2023-00259099, 100\%) supervised by the IITP (Institute for Information \& Communications Technology Planning \& Evaluation.)


\begin{thebibliography}{15}

\bibitem{batbaatar2025tl}
Batbaatar, E., Kim, J., Kim, Y., Yoon, Y.: Traversal Learning Coordination for Lossless and Efficient Distributed Learning. Expert Systems (2025). doi:10.1111/exsy.70141

\bibitem{mcmahan2017communication}
McMahan, H.B., Moore, E., Ramage, D., Hampson, S., Ag{\"u}era y Arcas, B.:
Communication-Efficient Learning of Deep Networks from Decentralized Data.
In: AISTATS, pp. 1273--1282 (2017)

\bibitem{bai2022constitutional}
Bai, Y., et al.: Constitutional AI: Harmlessness from AI Feedback.
arXiv:2212.08073 (2022)

\bibitem{christiano2017preferences}
Christiano, P.F., Leike, J., Brown, T.B., Martic, M., Legg, S., Amodei, D.:
Deep Reinforcement Learning from Human Preferences.
In: Advances in Neural Information Processing Systems 30 (2017)

\bibitem{yang2020pcpo}
Yang, T.-Y., Rosca, J., Narasimhan, K., Ramadge, P.J.:
Projection-Based Constrained Policy Optimization.
CoRR abs/2010.03152 (2020)

\bibitem{yu2020pcgrad}
Yu, T., Kumar, S., Gupta, A., Levine, S., Hausman, K., Finn, C.:
Gradient Surgery for Multi-Task Learning.
In: Advances in Neural Information Processing Systems 33, pp. 5824--5836 (2020)

\bibitem{shapley1953value}
Shapley, L.S.: A Value for n-Person Games. In: Contributions to the Theory of Games II, pp. 307--317. Princeton University Press (1953)

\bibitem{ghorbani2019data}
Ghorbani, A., Zou, J.:
Data Shapley: Equitable Valuation of Data for Machine Learning.
In: International Conference on Machine Learning, pp. 2242--2251 (2019)

\bibitem{koh2017understanding}
Koh, P.W., Liang, P.: Understanding Black-box Predictions via Influence Functions. In: ICML, pp. 1885--1894 (2017)

\bibitem{chen2024flce}
Chen, Y., Li, K., Li, G., Wang, Y.: Contributions Estimation in Federated Learning: A Comprehensive Experimental Evaluation. Proc. VLDB Endow. 17(8), 2077--2090 (2024)

\bibitem{murhekar2023incentives}
Murhekar, A., Yuan, Z., Chaudhury, B.R., Li, B., Mehta, R.: Incentives in Federated Learning: Equilibria, Dynamics, and Mechanisms for Welfare Maximization. In: NeurIPS, pp. 17811--17831 (2023)

\bibitem{luo2023pgfed}
Luo, J., Mendieta, M., Chen, C., Wu, S.: PGFed: Personalize Each Client's Global Objective for Federated Learning. In: ICCV, pp. 3923--3933 (2023)

\bibitem{sorensen2024pluralistic}
Sorensen, T., Moore, J., Fisher, J., Gordon, M.L., Mireshghallah, N., Rytting, C.M., Ye, A., Jiang, L., Lu, X., Dziri, N., Althoff, T., Choi, Y.: Position: A Roadmap to Pluralistic Alignment. In: ICML. PMLR 235, 46280--46302 (2024)

\bibitem{vepakomma2018split}
Vepakomma, P., Gupta, O., Swedish, T., Raskar, R.: Split Learning for Health: Distributed Deep Learning without Sharing Raw Patient Data. arXiv:1812.00564 (2018)

\bibitem{oh2024dpcutmixsl}
Oh, S., Baek, S., Park, J., Nam, H., Vepakomma, P., Raskar, R., Bennis, M., Kim, S.-L.: Privacy-Preserving Split Learning with Vision Transformers using Patch-Wise Random and Noisy CutMix. Trans. Mach. Learn. Res. (2024)

\bibitem{batbaatar2026tlaccuracyprivacypreserving}
Batbaatar, E., Yoon, Y.: TL++: Accuracy and Privacy Preserving Traversal Learning for Distributed Intelligent Systems. arXiv preprint arXiv:2606.25627 (2026). Available at: \url{https://arxiv.org/abs/2606.25627}

\end{thebibliography}
\end{document}